\documentclass[conference]{IEEEtran}
\IEEEoverridecommandlockouts
\usepackage{cite}
\usepackage{amsmath,amssymb,amsfonts}
\usepackage{algorithmic}
\usepackage{graphicx}
\usepackage{textcomp}
\usepackage{xcolor}
\usepackage{tikz}
\usepackage{enumitem}
\usepackage{colortbl}
\usepackage{booktabs}
\usepackage[hidelinks]{hyperref}
\usepackage{cleveref}

\def\BibTeX{{\rm B\kern-.05em{\sc i\kern-.025em b}\kern-.08em
    T\kern-.1667em\lower.7ex\hbox{E}\kern-.125emX}}
\begin{document}

\title{LOSTIN: \underline{L}ogic \underline{O}ptimization via \underline{S}patio-\underline{T}emporal \underline{In}formation with Hybrid Graph Models}

\author{ \normalsize Nan Wu$^1$, Jiwon Lee$^2$, Yuan Xie$^1$, and Cong Hao$^2$ \\
$^1$University of California - Santa Barbara, CA, USA \\
$^2$Georgia Institute of Technology, GA, USA \\
\{nanwu, yuanxie\}@ucsb.edu, \{jlee3251, callie.hao\}@gatech.edu \\}

\maketitle

\begin{abstract}
Despite the recent progress on machine learning (ML) based performance modeling, two major concerns that may impede production-ready ML applications in electronic design automation (EDA) are the stringent accuracy requirements and the generalization capability. 
To address these challenges, we a propose novel approach, namely LOSTIN\footnote{LOSTIN.com is a travel guide service to help people who visit a new city stay away from tourist traps and have a high-quality city tour. We envision our proposed LOSTIN would help the ML-based logic synthesis achieve high quality-of-results (QoR).}, 
which exploits hybrid graph neural networks (GNNs) to provide highly accurate quality-of-result (QoR) estimations with great generalization capability, specifically targeting logic synthesis optimization.
The key idea is to simultaneously leverage spatio-temporal information from hardware designs and logic synthesis flows to forecast performance (i.e., delay/area) of various synthesis flows on different designs.
Specifically, the structural characteristics inside hardware designs are distilled and represented by GNNs; 
the temporal knowledge (i.e., the relative ordering of logic transformations) in synthesis flows can be imposed on hardware designs by combining a virtually added supernode or a sequence processing model with conventional GNN models.
Evaluation on 3.3 million data points shows that the testing mean absolute percentage error (MAPE) on designs seen and unseen during training are no more than 1.2\% and 3.1\%, respectively, which are 7-15$\times$ lower than existing studies.
Our dataset and ML models are publicly available at
\textbf{\textcolor{teal}{\url{https://github.com/lydiawunan/LOSTIN}}}.
\end{abstract}

\begin{IEEEkeywords}
machine learning for electronic design automation, graph neural networks, logic synthesis
\end{IEEEkeywords}

\section{Introduction}

Despite the great advance achieved by electronic design automation (EDA) tools, there is still a long way towards \textit{hardware agile development}, whose ultimate goal is to reduce chip development cycles from years to months or even weeks.
To enable rapid optimization-evaluation iterations, one mainstay is to evaluate quality-of-results (QoRs) quickly and accurately.
Traditional EDA tools provide accurate estimations of QoR, yet are often time-consuming and require intensive manual efforts, leading to limited design space exploration and long time-to-market.
This situation is further exacerbated by the explosion of modern hardware system complexity and technology scaling.
Motivated by the strong desire for hardware agile development, a sound solution to fast and accurate QoR predictions is highly expected.

Recent years have witnessed the naissance of machine learning (ML) applied for computer systems and EDA tasks \cite{wu2022survey}, where one promising direction is ML-enabled fast performance modeling. 
Industrial investigations \cite{siemens} highlighted two basic requirements for production-ready ML in EDA: \textcircled{\small{1}} the \textit{accuracy} of ML-based performance estimation should be a minimum of $2 \sigma$ ($\sim $95\%); \textcircled{\small{2}} the \textit{generalization} capability is important for production-ready ML applications, meaning that an ML model should be able to directly apply to new designs without retraining.

\begin{figure}
    \centering
    \includegraphics[width=\linewidth]{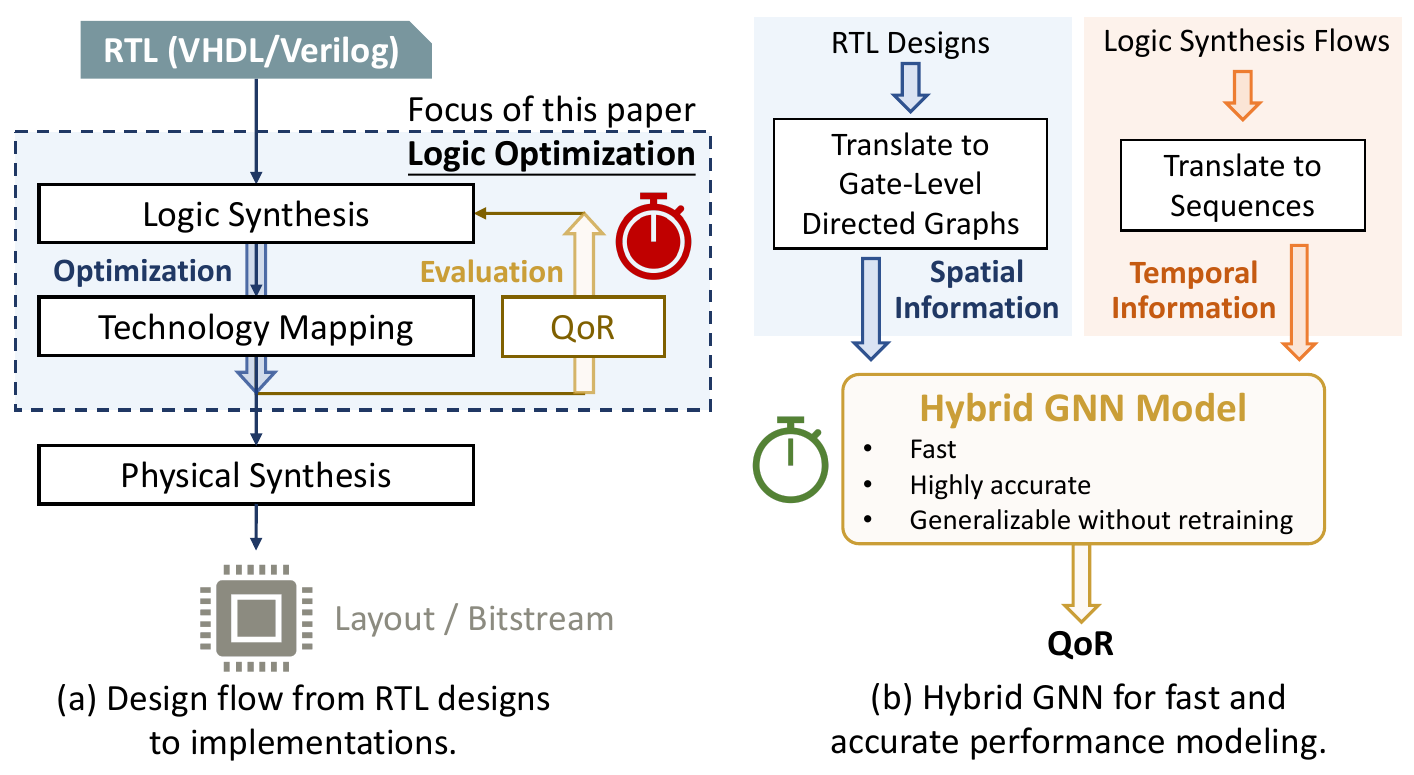}
    \caption{The EDA design flow and the proposed approach to predicting QoR after applying logic synthesis flows on hardware designs. 
    (a) The focus of this paper is to accelerate the evaluation phase in logic optimization.
    (b) The proposed model exploits spatial information from circuit designs and temporal knowledge from synthesis flows, generalizable to new designs without invoking any retraining.}
    \label{fig:overall1}
\end{figure}

\begin{figure*}[t]
    \centering
    \includegraphics[width=\textwidth]{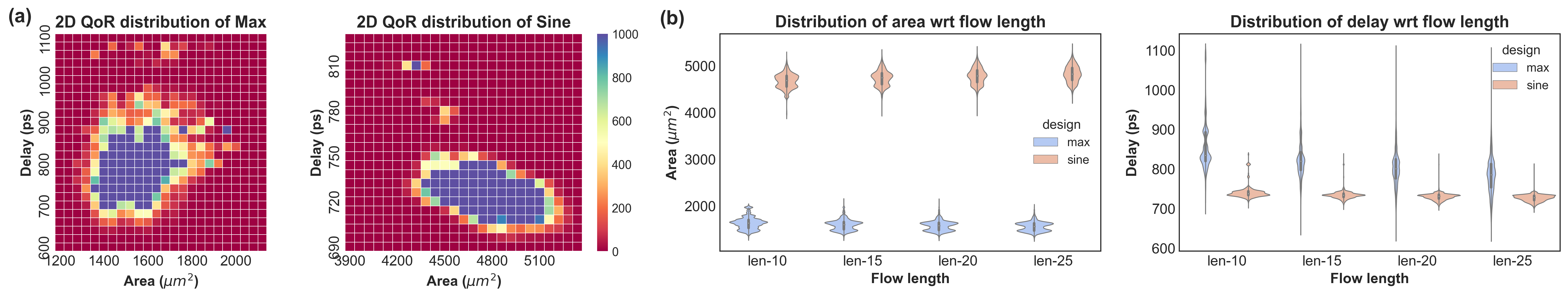}
    \caption{Area and delay results of 300,000 random ABC synthesis flows applied on \texttt{max} and \texttt{sin}, respectively. The number of count no less than 1,000 is represented by the same color.}
    \vspace{-10pt}
    \label{fig:motivation}
\end{figure*}

To this end, we target logic synthesis optimization, and propose a novel ML approach called LOSTIN for highly accurate QoR estimations with great generalization capability, as highlighted in Figure \ref{fig:overall1}.
Logic synthesis transforms functional RTL designs into optimized logic-gate-level representations.
A logic synthesis \textit{flow} refers to a sequence of logic optimizations, and a well designed flow can largely reduce design area and latency.
While being studied for decades, there remain unresolved \textbf{challenges and requirements} for efficient logic optimization, as follows.

\begin{itemize}
    \item The design space of possible synthesis flows is extremely large \cite{yu2018developing,yu2020decision}. It reemphasizes the importance of \textbf{fast and accurate} QoR prediction for sufficient design space exploration.
    \item There is no one-for-all solution. 
    Commercial EDA tools usually provide reference design flows \cite{lynx} developed by experts based on heuristics or prior knowledge, but such flows do not uniformly perform well.
    As shown in Figure \ref{fig:motivation}(a), first, for a specific design, different flows have drastically varied optimization effects; second, the same set of flows have different performance across different designs.
    These observations suggest the importance of \textbf{design-specific} synthesis flows.
    \item The \textit{transformation order} in the flows should be well captured.
    Figure \ref{fig:motivation}(b) compares the impact of different flow lengths, where the distribution of area/delay is not conspicuously improved with longer flows.
    This indicates that it is the underlying \textbf{temporal information}, i.e., the relative ordering of logic transformations, inside synthesis flows that majorly determines final QoRs.
    \item Existing approaches do not generalize across designs.
    Prior arts leverage CNN \cite{yu2018developing} or LSTM \cite{yu2020decision} to predict QoRs \textit{for a certain design}, where only flow-related but not design information is taken as input.
    These methods target fixed-length flows and have limited generalization capability to unseen designs due to the absence of design-specific information (more details in Section \ref{exp}).
    Aiming at a practical use of ML-based performance modeling, the \textbf{generalization} across different designs and flow lengths is a necessity.
\end{itemize}

Reckoning on the aforementioned issues, we innovatively emphasize utilizing the \textbf{spatio-temporal information} from both hardware designs and synthesis flows, upon which a hybrid graph neural network (GNN) based model is proposed for logic synthesis flow QoR prediction.
Specifically, the structural characteristics of hardware designs are distilled and represented by GNNs; the temporal knowledge among synthesis flows is extracted by a sequence processing model.
Equipped with the spatio-temporal knowledge, the proposed approach generalizes well on unseen designs.
We summarize our contributions as follows.
\begin{itemize}
    \item \textbf{Modeling}. We propose two generalizable GNN-based approaches to predict the performance of logic synthesis flows, which incorporate crucial spatio-temporal information from both hardware designs and synthesis flows.
    The first model utilizes a supernode on GNN to encode the synthesis flow impact on the circuit. 
    The second model combines LSTM and GNN in a hybrid manner, to capture the temporal knowledge of flows.
    In particular, the second model \textit{represents the graphs and synthesis flows separately}, which significantly reduces the training complexity and memory overhead; it also better represents the problem nature, where circuits (represented by graphs) and synthesis flows (represented by sequence) are two separate concepts. 
    \item \textbf{Evaluation}. Evaluations on both \textit{seen and unseen} (during training) designs demonstrate the superiority of our approach.
    On seen designs, i.e., transductive scenario, the mean absolute percentage error (MAPE) achieved by the hybrid GNN is less than 1.2\%, which is 7$\times$ lower than existing works.
    On unseen designs, i.e., inductive scenario, the MAPE is still below 3.2\%, $14\times$ lower than existing works. It demonstrates the extraordinary generalization capability across designs with zero retraining.
    \item \textbf{Insights}. We provide insights on graph representation learning in EDA problems. 
    \textcircled{\small{1}} With carefully selected GNN models, stacking more layers (i.e., deep GNN) introduces a performance boost in prediction accuracy.
    \textcircled{\small{2}} The temporal information from synthesis flows (not limited to logic synthesis) can be imposed on hardware netlists by combining a supernode or a sequence processing model with conventional GNN models.
    \item \textbf{Dataset}. We provide an open-source dataset consisting of 3.3 million data points collected from 11 different circuit designs, with the goal to facilitate multi-modal or dynamic graph representation learning for EDA tasks.
\end{itemize}


\section{Related Work and Preliminary}

\subsection{Related Work}
In logic optimization, the sequence to apply logic transformations, i.e., \textit{the logic synthesis flow}, is often determined heuristically.
For example, commercial EDA tools provide reference synthesis flows \cite{lynx}; 
an academia open-source logic synthesis tool ABC \cite{brayton2010abc} offers synthesis flows \textit{resyn}, \textit{resyn2} and \textit{resyn2rs}.

Recently, ML-assisted logic optimization has attracted increasing research interests, aiming to reduce exploration time and improve performance.
For example, LSOracle \cite{neto2019lsoracle} employs multi-layer perceptron (MLP) to automatically decide which one of the two optimizers should be applied on different parts of circuits.
The logic optimization can also be formulated as a reinforcement learning problem, implemented with a GNN-based agent \cite{haaswijk2018deep,zhu2020exploring} or a non-graph based agent \cite{hosny2020drills}.
The optimization objective is to minimize area  \cite{haaswijk2018deep,zhu2020exploring,hosny2020drills} or delay \cite{haaswijk2018deep}.
In terms of forecasting logic synthesis flow performance, 
Yu \textit{et al.} \cite{yu2018developing} use a convolution neural network (CNN) to identify whether a synthesis flow is an angel-flow or a devil-flow.
Later, they adopt a long short-term memory (LSTM) \cite{yu2020decision} network with transfer learning to predict delay and area after applying a synthesis flow.

From a broader view, GNNs are expected to make better use of graph structured data in many EDA problems \cite{khailany2020accelerating}.
Instead of conventional graph representation learning that maps circuit designs from static graphs to labels (e.g., resource/timing/power) \cite{zhang2020grannite,wu2022survey,wu2021ironman}, the target task in this work should consider both designs (i.e., static graphs) and synthesis flows (i.e., transformations to be applied on the graphs) to provide high-accuracy predictions of delay/area, which can be recognized as \textit{a multi-modal or dynamic graph representation learning}.

\subsection{Preliminary}
\textbf{GNN.}
GNNs operate by propagating information along the edges of a given graph.
Specifically, each node $v$ is initialized with an representation $h_v^{(0)}$, which could be a direct representation or a learnable embedding obtained from features of this node.
Then, a GNN layer updates each node representation by integrating the node representations of its neighbors in the graph and itself, yielding representations $h_v^{(1)}$.
This process can be unrolled through time steps by repeatedly using the same update function, deriving representations $h_v^{(2)}, h_v^{(3)}, ..., h_v^{(T)}$.
An alternative is to stack several GNN layers, intuitively similar to unrolling through time steps, but increases the GNN capacity by using different parameters in the update function for each time step.
Notably, the more GNN layers are stacked, the larger the receptive field is.
Several prevalent GNN examples are the graph convolutional network (GCN) \cite{kipf2016semi} and the graph isomorphism network (GIN) \cite{xu2018powerful}: 
GCN is inspired by the first-order graph Laplacian methods, and essentially performs aggregation and transformation on node representations without learning trainable filters;
GIN is provably as powerful as the Weisfeiler-Lehman graph isomorphism test, leveraging sum aggregators over a countable input feature space.

\textbf{Supernode in GNNs.} 
The introduction of a supernode aims to address the difficulty in propagating information across remote parts of graphs \cite{gilmer2017neural,ishiguro2019graph}.
The supernode is a newly added virtual node that connects all the nodes in the original graph to promote global information propagation by reducing the maximum distance between any two nodes to two hops.
Many GNN models can be equipped with such a supernode, which serves as a global scratch space that every other node reads from and writes to in every step of message passing with some preference.

\textbf{LSTM.}
Long short-term memory \cite{hochreiter1997long} network is a type of recurrent neural network capable of learning the order dependence and long-term dependence in sequence processing problems. 
For each unit in a sequence, the same computations are performed and the current output states are dependant on the previous (hidden) states.
Theoretically, LSTM can process sequential inputs such as sentences in arbitrary length.
A common LSTM unit consists of a cell, an input gate, an output gate, and a forget gate, among which the cell is responsible to remember values over arbitrary time intervals and the rest three gates regulate the information fed into and out of the cell.
Given its sequential information processing capability, an LSTM-based model is a proper candidate to represent synthesis flows.


\begin{figure*}[tp]
    \centering
    \includegraphics[width=\textwidth]{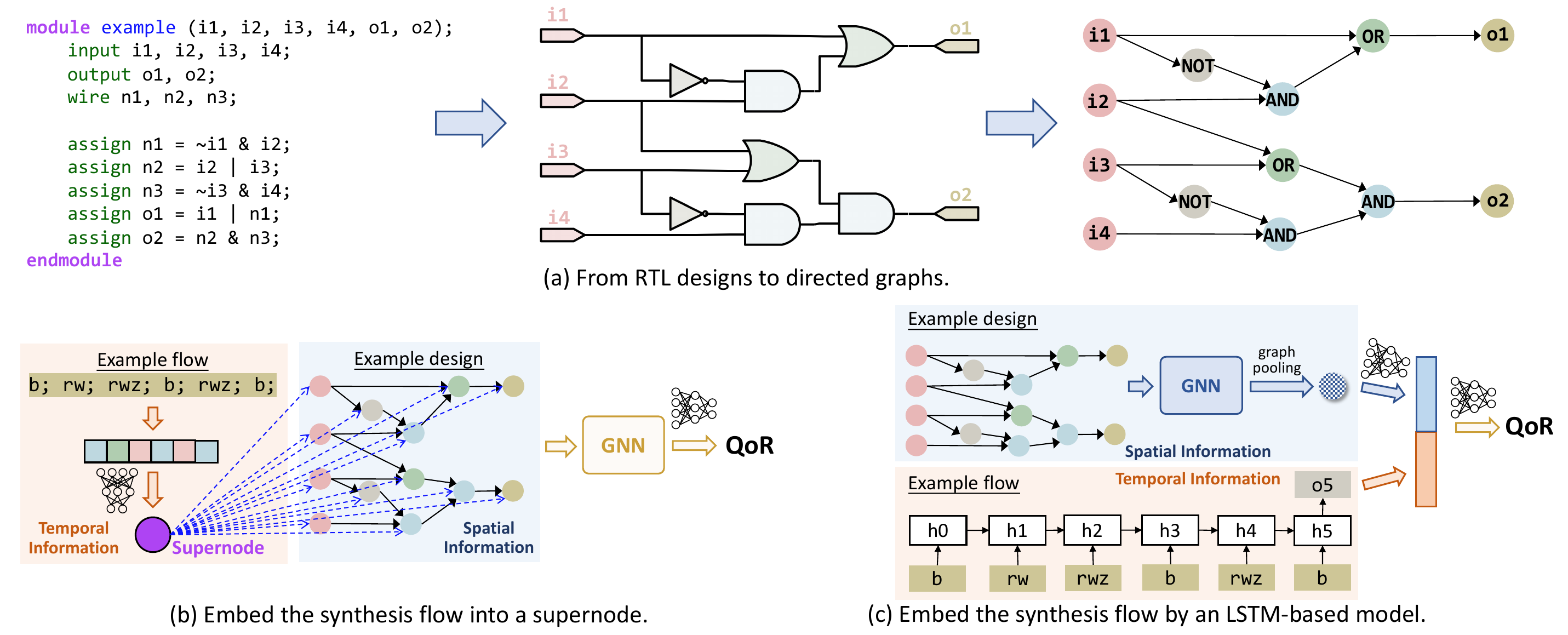}
    \vspace{-10pt}
    \caption{The overview of our proposed GNN architectures. (a) Logic synthesis takes in register-transfer-level (RTL) descriptions and converts to gate-level netlists, from which we build directed graphs. (b) The proposed GNN with supernode. (c) The proposed hybrid GNN with LSTM.}
    \vspace{-10pt}
    \label{fig:overall2}
\end{figure*}

\section{Proposed Hybrid Models}
We present LOSTIN, a novel, fast, accurate, and generalizable ML approach for QoR estimations of logic synthesis flows, exploiting spatio-temporal information.
Two models are explored: 
\textcircled{\small{1}} a GNN for spatial information learning, armed with a supernode to encode temporal information (Section \ref{sec:gnn-virtual});
\textcircled{\small{2}} a hybrid model, composed of a GNN for spatial learning and an LSTM for temporal learning (Section \ref{sec:gnn-hybrid}).

\subsection{Problem Formulation}\label{sec:formulation}

In logic synthesis, hardware designs are converted to logic networks, which are typically graph abstractions of logic circuit implementations in the gate level.
Logic optimization aims to manipulate and transform logic networks to reduce the amount of required hardware or the critical path delay by sequences of logic transformations, which are referred to as logic synthesis flows.

\textbf{Prediction task.}
We leverage ABC~\cite{brayton2010abc}, an open-source logic synthesis framework well-adopted in academia, to produce synthesis flows.
Notably, the adopted ABC tool can be arbitrarily changed to other synthesis tools as long as enough training data are collected.
The inputs to the proposed predictors are initial hardware designs described in RTL and synthesis flows to be applied.
The QoR metrics to be predicted are the logic area (denoted as area) and the critical path delay (denoted as delay). 
The ground truth is collected from ABC after technology mapping.
Further, the formulation of this prediction task can be extended to other flow performance estimation problems, such the resource utilization in high-level synthesis \cite{wu2022high} and negative slacks in placement and routing (i.e., physical synthesis)\cite{pan200821,kahng2018using}.

\textbf{Graph representation for circuits.}
As logic optimization targets gate-level transformations, we represent circuit designs as directed graphs, where each node is a primary logic gate and each edge shows logic dependency.
A Verilog parser is built to translate RTL designs into gate-level netlists.
To guarantee universal representations of \textit{any combinational logic functions}, \texttt{AND}, \texttt{OR}, and \texttt{NOT} gates are included in the translated graphs.
Multiple-output or more-than-two-input gates are automatically split and parsed into the three types of gates aforementioned.
Thus, each node has two attributes: \textcircled{\small{1}} node type in  input/intermediate/output, and \textcircled{\small{2}} operation type in \texttt{AND}/\texttt{OR}/\texttt{NOT}.
Such a parser enables the circuit representations to be independent of logic optimizers targeting different logic representations \cite{neto2019lsoracle} (e.g., And-Inverter Graphs (AIGs) \cite{brayton2010abc} and Majority-Inverter Graphs (MIGs) \cite{amaru2014majority}), fostering the portability across different logic synthesis tools.
The process of transforming an RTL design into a directed graph is exemplified in Figure \ref{fig:overall2}(a).

\textbf{Flow representation.}
Within the ABC framework, we consider synthesis flows composed of 7 types of logic transformation actions from $\mathbb{S} = \{$\textit{balance (b), resubstitution (rs), resubstitution -z (rsz), rewrite (rw), rewrite -z (rwz), refactor (rf), refactor -z (rfz)}$\}$.
To integrate the inherent temporal information from synthesis flows with circuit designs, a synthesis flow can be represented as either \textcircled{\small{1}} a vector to construct a supernode that directly propagates temporal knowledge to circuit designs (Section \ref{sec:gnn-virtual}) or \textcircled{\small{2}} a sequence embedding generated by an LSTM model (Section \ref{sec:gnn-hybrid}).

\textbf{Efficiency of separate encoding of circuits and sequences, compared with the unified encoding.}
The most naive way is to merge a circuit design with one particular synthesis flow to generate a single graph (where the flow becomes node attributes), and directly apply vanilla GNN models to produce unified representations.
However, this approach has two major issues.
First, by plugging the flow into node attributes, all the nodes will share similar node representations, resulting in over-redundant input features.
Second, it is nearly impossible to build a graph dataset, in which each graph is large-scale and each node has many node attributes, for graph-level regression.
Current graph representation learning pays attention to either graph-level tasks on relatively small graphs \cite{maron2019provably} or node/link-level tasks on large graphs \cite{fan2019graph}.
Given a considerate number of large graphs multiplying copious node attributes, the dataset generation process causes out-of-memory issues, impractical for implementation and vulnerable to scalability.
In our preliminary experiments, such a unified representation would generate a dataset over 80 gigabytes, which is challenging for efficient training.
Instead, we propose to represent the circuit designs and synthesis flows separately, so that they can be processed and stored separately; during training, the graph representation does not need to be repeatedly loaded for different synthesis flows. 
Such a separate approach significantly reduces the memory overhead for storage and training, as well as the training time by orders of magnitude.

\subsection{GNN with Supernode}
\label{sec:gnn-virtual}

Inspired by the idea that introducing a supernode to graphs can collect and redistribute global information with some preference \cite{gilmer2017neural,ishiguro2019graph}, we propose to leverage a supernode to represent synthesis flows.
Since the supernode is virtually connected to all the nodes in the original graph, the temporal information is directly injected into the circuit graph, as shown in Figure~\ref{fig:overall2}(b).

\textbf{Supernode to embed synthesis flows}.
Synthesis flows are converted to fixed-length input vectors with the dimension of 25, since the maximum length of currently considered synthesis flows is 25.
Each logic transformation in a flow is represented as an integer from 1 to 7, and zero padding is applied for flows shorter than 25.
In Figure \ref{fig:overall2}(b), the initial input vector of the example flow is encoded as [1, 6, 7, 1, 7, 1, 0, ...].
A single fully-connected layer then converts the $1\times25$ vector into $1\times 8$ as the supernode embedding.

\textbf{Spatial representation of circuit structures}.
To study the impact of synthesis flows on a circuit design, we connect the supernode to all other nodes in the original graph.
The two node attributes are converted to learnable node embeddings as $1 \times 8$ vectors.
The modified graphs are passed through GNN models for graph representation learning.
By exposing the temporal information encoded in the supernode and distributing it to the entire graph, this model is expected to process both spatial and temporal features simultaneously, i.e., learn the effects of synthesis flows on different circuit structures.

\subsection{Hybrid GNN with Spatio-Temporal Information} 
\label{sec:gnn-hybrid}

While a supernode is capable to collect global information and distribute temporal knowledge to every other node in graphs, we notice three concerns that may influence prediction performance.
First, synthesis flows are represented by fixed-length vectors, which are then passed through an MLP to comply with the embedding dimension of other nodes.
This setting is insensitive to sequence dependence, i.e., the transformation order in synthesis flows, whereas the impact of later transformations heavily depends on previous ones.
Second, as the message passing process proceeds, the original temporal information inside the supernode is gradually faded in other nodes.
Third, by adding the supernode that connects all the nodes in the original graph, the graph size increases with newly added edges, which may cause scalability issue in implementation when encountering extremely large graphs.

To address the first concern, a more natural way is to leverage a sequence processing model to distill the temporal information, and the specific model employed in this work is LSTM, which excels at handling order dependence and variable-length flows.
To resolve the second and the third concerns, we separately generate a sequence embedding (i.e., a synthesis flow representation) and a graph embedding (i.e., a circuit representation) in the feature extraction stage, and these two embeddings are concatenated for downstream predictions.
The approach of separately learning two embeddings and then concatenating is intuitively consistent with the actual logic synthesis procedure, since it mimics the operation of applying the synthesis flow to the circuit.

Figure \ref{fig:overall2}(c) illustrates the structure of the second hybrid GNN model with LSTM.
The directed graph translated from a circuit design is passed through a GNN model, followed by a linear layer to generate a graph-level representation (i.e., a $1 \times 32$ vector).
The synthesis flow is processed by a two-layer LSTM to derive a flow representation (i.e., a $1\times 64$ vector).
These two vectors are concatenated to form a $1\times 96$ vector.
Finally, a feed-forward MLP with the structure of 96-100-100-1 is adopted for delay/area predictions.
\section{Experiment}
In this section, we first describe the dataset generation and the setup for our experiments, and then present our evaluations with discussions on the results.

\subsection{Dataset Generation}
We select eleven circuit designs from the EPFL benchmark \cite{amaru2015epfl}, a benchmark suite designed as a comparative standard for logic optimization and synthesis.
The logic synthesis flows are generated by the logic synthesis tool ABC \cite{brayton2010abc}.
To demonstrate the flexibility of handling variable-length synthesis flows, we create synthesis flows consisted of 10, 15, 20, and 25 logic transformations;
for each different length, there are 50,000, 50,000, 100,000, and 100,000 flows, respectively, totally making up 300,000 flows.
Each synthesis flow consists of logic transformations from $\mathbb{S} = \{$\textit{b, rs, rsz, rw, rwz, rf, rfz}$\}$.
All the 300,000 flows are applied to eleven different designs with ASAP 7nm low-voltage technologies \cite{xu2017standard}, which are 3.3 million data points in total.
The ground truth (i.e., label) is collected from ABC after technology mapping.

\subsection{Experimental Setup}

%

\textbf{Baseline.}
The proposed hybrid GNN models are compared against two existing ML-based approaches: a CNN-based model \cite{yu2018developing}, and an LSTM-based model \cite{yu2020decision}.
We exactly follow the model structures mentioned in the prior work, with minor modifications to fit our prediction task.
\begin{itemize}
    \item In the CNN baseline, each transformation is represented as a one-hot vector; synthesis flows (with the maximum length of 25) are represented as $7\times 25$ matrices with zero padding for shorter flows. 
    We honor the original CNN structure \cite{yu2018developing}, in which there are two convolutional layers followed by a max-pooing and two FC layers. 
    Since the CNN-based model was designed for binary classification \cite{yu2018developing}, we replace the final classifier by a single neuron for value prediction in our task. 
    Note that the CNN-based model can only be trained in a \textit{design-specific} manner, i.e., one model for one design.
    
    \item In the LSTM baseline, we change the one-hot embeddings of transformations to learnable embeddings. 
    Similar to the CNN-based model, although the LSTM-based model should also be design-specific, we intentionally study its generalizability by training one model for all. 
    To distinguish different designs, we add design names as the identity preceding to synthesis flows to construct new sequences.
    This model has two LSTM layers with the hidden size of 128, followed by an MLP of 128-30-30-1 to predict delay/area.
\end{itemize}

\begin{table}[tb]
\caption{Graph size of different circuit designs.}
\vspace{-5pt}
\label{graphsize}
\centering
    \renewcommand{\arraystretch}{1}
    \setlength{\tabcolsep}{16pt}
\begin{tabular}{c|c|c}
\toprule
 & \textbf{Number of Nodes} & \textbf{Number of Edges} \\ \midrule
\textbf{adder}   & 2926 & 3690 \\
\textbf{arbiter} & 24258 & 35841 \\
\textbf{bar} & 6935 & 10136 \\
\textbf{div} & 143375 & 200494 \\
\textbf{log2} & 68881 & 100909 \\
\textbf{max} & 6752 & 9105 \\
\textbf{multiplier} & 59404 & 86338 \\
\textbf{sin} & 11486 & 16878 \\
\textbf{sqrt} & 57296 & 81786 \\
\textbf{square} & 42042 & 60460 \\
\textbf{voter} & 35105 & 47862 \\
\bottomrule
\end{tabular}
\vspace{-10pt}
\end{table}

\textbf{Implementation and training.}
All the aforementioned neural network models are implemented with Pytorch \cite{paszke2019pytorch} and Pytorch Geometric \cite{Fey/Lenssen/2019}.
Experiments were performed on a Linux host with a 64-core Intel Xeon Gold 5218 CPU (2.30 GHz) and Nvidia RTX 2080Ti GPUs.

Training, validation, and testing sets are split by 20:5:75. 
We highlight two training and evaluation strategies.
\underline{First}, in contrast to many ML tasks that use a large proportion of the entire dataset for training, we intentionally train the proposed models with a small portion and conduct evaluations on the rest data points. 
\underline{Second}, we evaluate both \textbf{transductive and inductive} scenarios.
If a design is seen during training but with unseen flows in the testing, it is referred to as transductive; 
if a design is unseen during training and only employed for testing, it is referred to as inductive.
The goal is to emphasize the generalizability of our proposed models, which is important for many EDA tasks that are possibly suffering from data scarcity.
Among eleven circuit designs, six of them (\texttt{adder}, \texttt{arbiter}, \texttt{bar}, \texttt{div}, \texttt{log2}, and \texttt{max}) are used for both training and testing, and the remaining five (\texttt{multiplier}, \texttt{sin}, \texttt{sqrt}, \texttt{square}, and \texttt{voter}) are \textbf{merely evaluated in testing to demonstrate generalization, i.e., inductive capability}. Training details are summarized as follows. 

\begin{itemize}
    \item For every design, we train a design-specific CNN for 20 epochs with the RMSprop optimizer (learning rate 0.05).
    \item LSTM is trained for 100 epochs with the Adam optimizer (initial learning rate 2e-3, weight decay 2e-6).
    \item For the GNN with supernode (denoted as GNN-S), a ten-layer GIN model is trained for 20 epochs with the Adam optimizer (learning rate 1e-3); node and edge embedding dimensions are 8 and 2, respectively. 
    \item For the hybrid GNN-LSTM model (denoted as GNN-H), a ten-layer GIN is combined with a two-layer LSTM (whose hidden size is 64), trained for 20 epochs with the Adam optimizer (initial learning rate 2e-3, weight decay 2e-6). The node embedding dimension is 32. 
\end{itemize}

\subsection{Evaluation}
\label{exp}

\textbf{Transductive scenario}.
Table \ref{table1} shows the MAPE of QoR predictions on the designs that are seen during training but with unseen synthesis flows.
We have the following observations.
\textcircled{\small{1}} Since the CNN baseline is design-specific, it slightly outperforms the LSTM-based model, which is a unified model across all designs.
\textcircled{\small{2}} The hybrid GNN model, GNN-H, significantly outperforms the LSTM-based model, with $7\times$ and $15\times$ lower MAPE than those of area and delay prediction, respectively. 
\textcircled{\small{3}} The GNN with supernode, GNN-S, shows comparable performance with the LSTM-based model.

\textbf{Inductive scenario}.
Table \ref{table2} shows the MAPE of QoR predictions on unseen designs.
\textcircled{\small{1}} The CNN-based model only works for design-specific synthesis flows and thus there is no generalization to unseen designs.
\textcircled{\small{2}} The LSTM-based model suffers from a large accuracy degradation for unseen designs, indicating limited generalization capability.
\textcircled{\small{3}} The GNN-S slightly outperforms the LSTM-based model by 3\% and 9\% in area and delay prediction, respectively (further discussed in Section \ref{discussion}).
\textcircled{\small{4}} The GNN-H maintains its high prediction accuracy by slightly increasing the MAPE from $1\%$ to $3\%$, demonstrating extraordinary generalization capability.

\begin{figure}[t]
    \centering
    \includegraphics[width=\linewidth]{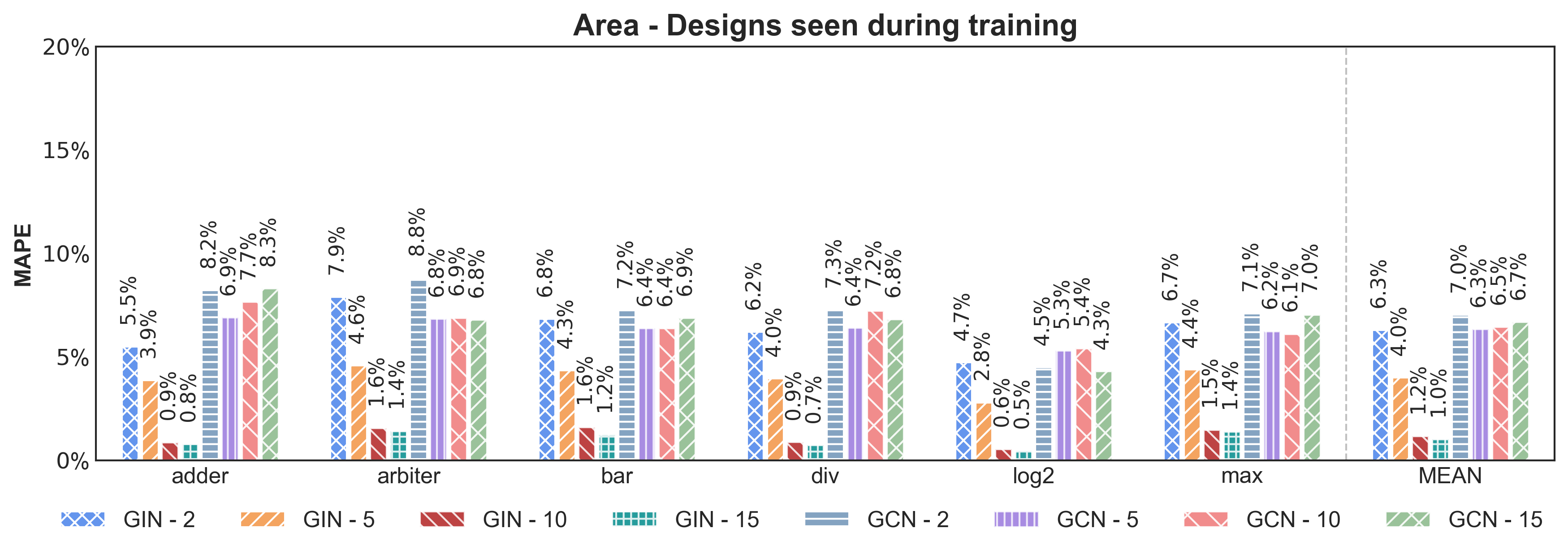}
    \caption{Transductive MAPE on area predictions made by GNN-H. Results are compared in terms of GIN and GCN with different number of layers.}
    \label{fig:gnn_layer_area_seen}
\end{figure}

\begin{figure}[t]
    \centering
    \includegraphics[width=\linewidth]{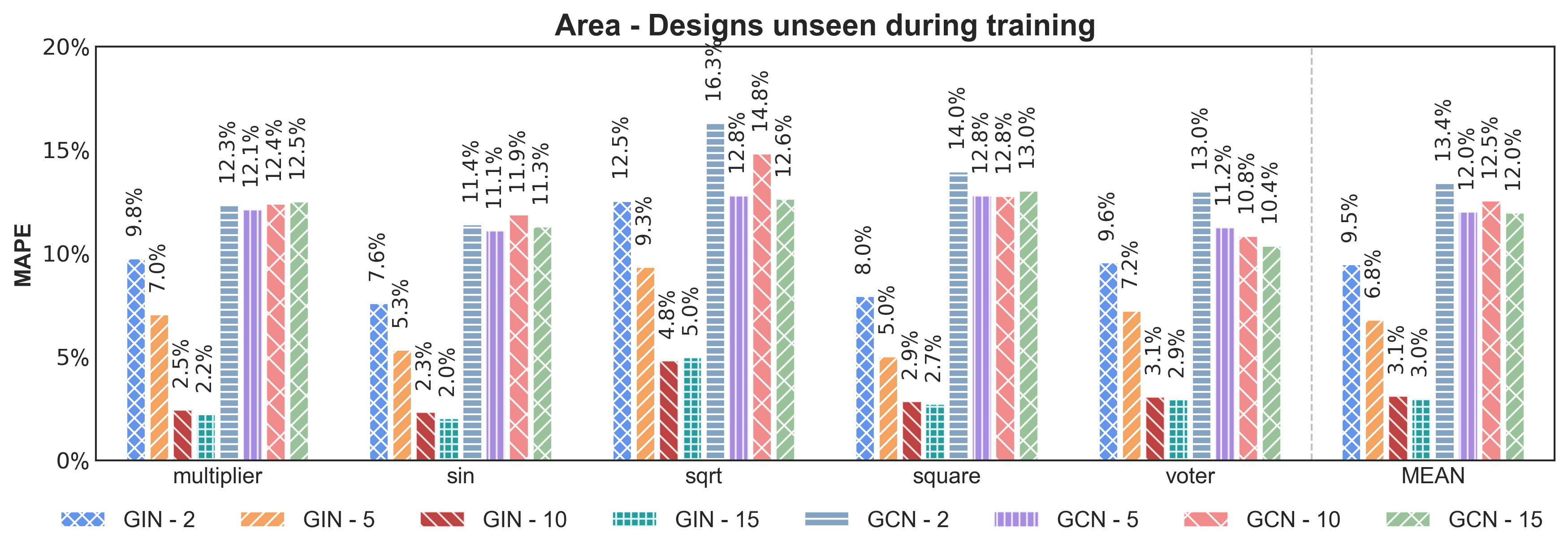}
    \caption{Inductive MAPE on area predictions made by GNN-H. Results are compared in terms of GIN and GCN with different number of layers.}
    \label{fig:gnn_layer_area_unseen}
    \vspace{-10pt}
\end{figure}

\begin{figure}[t]
    \centering
    \includegraphics[width=\linewidth]{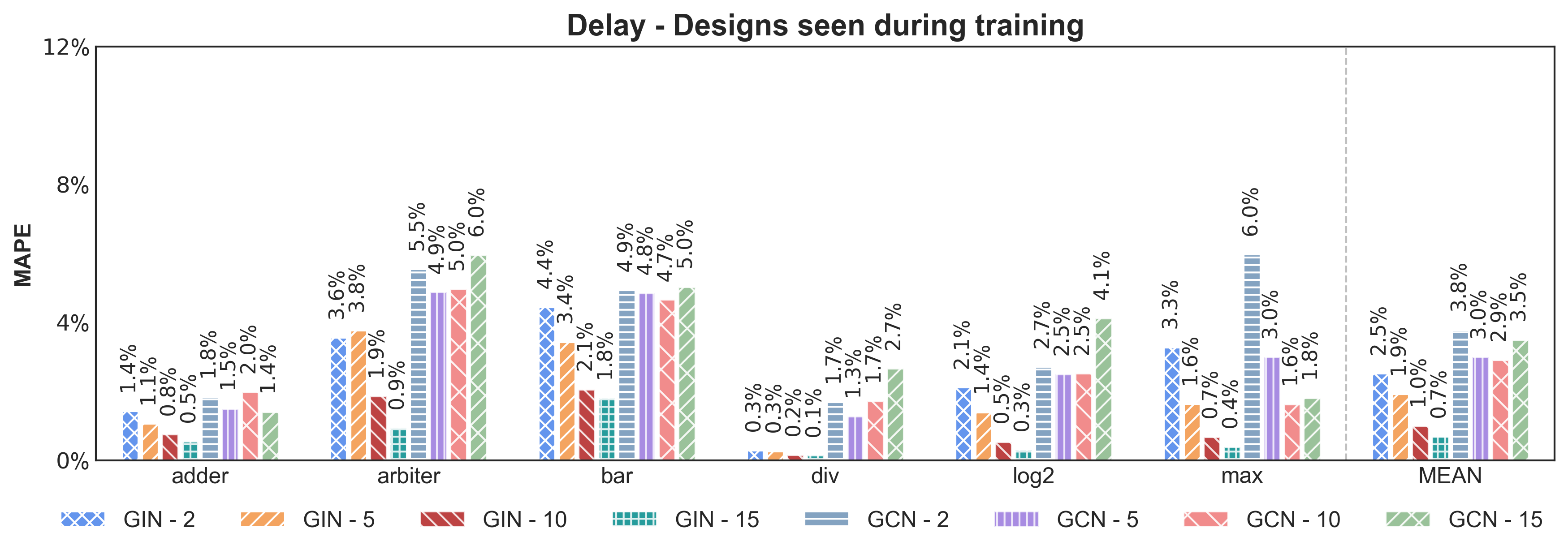}
    \caption{Transductive MAPE on delay predictions made by GNN-H. Results are compared in terms of GIN and GCN with different number of layers.}
    \label{fig:gnn_layer_delay_seen}
\end{figure}

\begin{figure}[t]
    \centering
    \includegraphics[width=\linewidth]{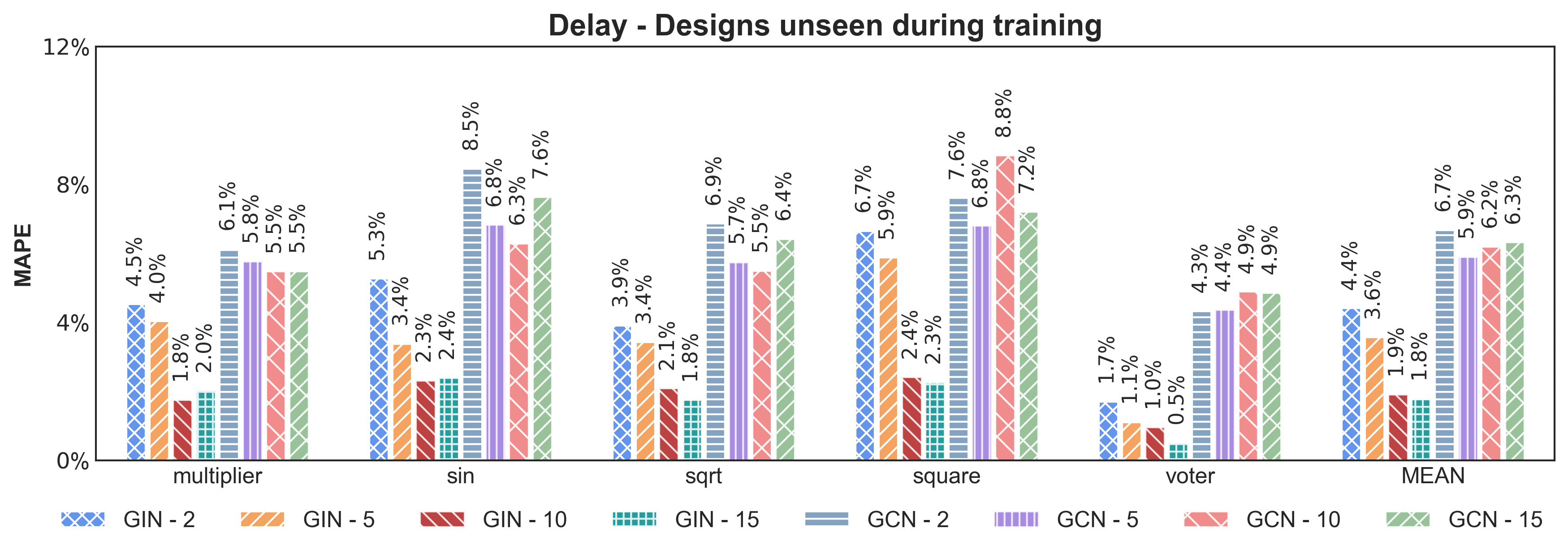}
    \caption{Inductive MAPE on delay predictions made by GNN-H. Results are compared in terms of GIN and GCN with different number of layers.}
    \label{fig:gnn_layer_delay_unseen}
    \vspace{-10pt}
\end{figure}

\textbf{Sensitivity analysis.}
We study the design choices of GNN-H in terms of GNN types and the number of layers.
Figure~\ref{fig:gnn_layer_area_seen}-\ref{fig:gnn_layer_delay_unseen} compare the MAPE of QoR predictions with respect to both GIN and GCN models with different number of layers.
Generally, GIN models receive an accuracy boost after stacking ten layers, whereas GCN models show similar prediction accuracy among different choices of layers.
\textcircled{\small{1}} 
Regarding the GNN type comparison, GCN suffers from the over-smoothing problem \cite{liu2020towards}.
Mathematically, GCN~\cite{kipf2016semi} is an approximate of $2I_N - L$, where $L$ is the normalized graph Laplacian operator and $I_N$ is the identity matrix.
Since the graph Laplacian operator/filter is a high-pass filter, GCN naturally becomes a low-pass filter, indicating that stacking many layers does not help better characterize graph structures.
\textcircled{\small{2}}
Regarding the number of GNN layers, a deep GNN setting with carefully selected GNN types possesses better representation power, since stacking more layers brings a larger receptive field to characterize input graphs and provides hierarchical abstractions of input structures, especially beneficial for large graphs.

\begin{table}[tp]
\vspace{-5pt}
\caption{Comparison with CNN~\cite{yu2018developing} and LSTM~\cite{yu2020decision} in the \textbf{TRANSDUCTIVE} scenario. \textbf{GNN-S} is the proposed GNN with supernode; \textbf{GNN-H} is the proposed hybrid GNN.}
\label{table1}
\centering
    \renewcommand{\arraystretch}{1}
    \setlength{\tabcolsep}{1.5pt}
\begin{tabular}{c|cccc|cccc}
\toprule
\textbf{} & \multicolumn{4}{c|}{\textbf{Area (MAPE)}} & \multicolumn{4}{c}{\textbf{Delay (MAPE)}} \\
\textbf{} & \textbf{CNN} & \textbf{LSTM} & \textbf{GNN-S} & \textbf{GNN-H} & \textbf{CNN} & \textbf{LSTM} & \textbf{GNN-S} & \textbf{GNN-H} \\ \midrule
\textbf{adder}   & 7.00\%  & 8.72\%  & 7.65\%  & 0.87\% & 1.76\% & 16.22\% & 1.79\%  & 0.76\% \\
\textbf{arbiter} & 2.98\%  & 13.66\% & 8.16\%  & 1.56\% & 0.23\% & 18.96\% & 15.37\% & 1.86\% \\
\textbf{bar}     & 8.46\%  & 5.22\%  & 22.72\% & 1.61\% & 0.74\% & 14.98\% & 22.59\% & 2.06\% \\
\textbf{div}     & 12.71\% & 7.75\%  & 13.16\% & 0.88\% & 7.72\% & 14.31\% & 9.29\%  & 0.16\% \\
\textbf{log2}    & 8.04\%  & 9.05\%  & 4.19\%  & 0.55\% & 3.87\% & 11.85\% & 12.74\% & 0.53\% \\
\textbf{max}     & 7.28\%  & 6.18\%  & 8.35\%  & 1.48\% & 5.50\% & 17.37\% & 20.60\% & 0.68\% \\ \midrule
\textbf{MEAN}    & 7.75\%  & 8.43\%  & 10.70\% & 1.16\% & 3.30\% & 15.62\% & 13.73\% & 1.00\% \\ 
\bottomrule
\end{tabular}
\end{table}

\begin{table}[tp]
\caption{Comparison with LSTM~\cite{yu2020decision} in the \textbf{INDUCTIVE} scenario. }
\vspace{-5pt}
\label{table2}
\centering
    \renewcommand{\arraystretch}{1}
    \setlength{\tabcolsep}{4pt}
\begin{tabular}{c|ccc|ccc}
\toprule
\textbf{} & \multicolumn{3}{c|}{\textbf{Area (MAPE)}} & \multicolumn{3}{c}{\textbf{Delay (MAPE)}} \\
\textbf{}  & \textbf{LSTM} & \textbf{GNN-S} & \textbf{GNN-H} & \textbf{LSTM} & \textbf{GNN-S} & \textbf{GNN-H} \\ \midrule
\textbf{multiplier}   & 57.82\%  & 9.39\%  & 2.45\% &  38.21\% & 17.89\% & 1.75\% \\
\textbf{sin}          & 66.09\%  & 64.48\% & 2.34\% &  45.94\% & 54.44\% & 2.32\% \\
\textbf{sqrt}         & 29.03\%  & 39.25\% & 4.83\% &  38.03\% & 15.75\% & 2.09\% \\
\textbf{square}       & 38.59\%  & 13.96\% & 2.86\% &  47.52\% & 31.34\% & 2.41\% \\
\textbf{voter}        & 27.38\%  & 76.49\% & 3.08\% &  42.19\% & 46.54\% & 0.96\% \\ \midrule
\textbf{MEAN}         & 43.78\%  & 40.71\% & 3.11\% &  42.38\% & 33.20\% & 1.91\% \\ 
\bottomrule
\end{tabular}
\vspace{-10pt}
\end{table}

\subsection{Discussion and Insight}
\label{discussion}

\textbf{GNN-S v.s. GNN-H}.
In GNN-S, even though a synthesis flow is encoded as a supernode, there are several limitations that influence temporal information characterization.
First, every synthesis flow is directly represented as a fixed-length vector to generate a supernode embedding, which is insensitive to sequence dependence, i.e., the order of logic transformations.
Second, the original temporal information injected to the supernode is gradually diluted, since the supernode embedding also evolves during the message passing. 
We compared different communication mechanisms between the supernode and other nodes, i.e., bidirectional or unidirectional, where the prediction accuracy is slightly improved with the unidirectional communication. 
This demonstrates that the slower the dilution rate is, the more temporal knowledge can be reserved during learning.
Third, simply adding a supernode into original graphs may not be an efficient approach to fusing information from different modalities (i.e., graph-structured data and sequence-structured data).
By contrast, GNN-H leverages a more direct scheme that takes advantages of both GNN and LSTM to extract spatio-temporal information in a decoupled manner.
During the feature extraction, the LSTM directly characterizes temporal information from synthesis flows, and the GNN focuses on representing spatial structures of circuit designs.
Rather than mixing spatial and temporal information at the very first step (as in GNN-S), separately built and learned graph and sequence embeddings have better expressiveness for each source of input information, thus providing a better foundation for downstream tasks.

\textbf{Scalability regarding graph abstraction level.}
Table \ref{graphsize} shows the gate-level graph size of different circuit designs.
The bright side is both GNN-S and GNN-H can handle large graphs.
The dark side is the graph size will explode for larger circuit designs, which may cause scalability issues in practical implementation.
Regarding different GNN-based models, the vanilla GNNs, as aforementioned in Section \ref{sec:formulation}, cause out-of-memory issues even with the current dataset, letting alone larger circuit designs.
GNN-S may exhibit some scalability concerns, since it adds a considerable number of virtual edges to original graphs, i.e., $|V|$ virtual edges will be newly added for a graph original with $|V|$ nodes.
Two potential directions to further improve scalability are \textcircled{\small{1}} extracting graphs from higher level of circuit abstractions to provide graphs in proper sizes so that both GNN-S and GNN-H can easily handle, or \textcircled{\small{2}} clustering nodes in gate-level graphs hierarchically \cite{bateni2017affinity} to guarantee reasonable compute cost for each stage.

\textbf{Multi-modality graph representation learning.}
Graph representation learning has evolved from single-modal to multi-modal \cite{holzinger2021towards}, with several attempts of exploiting multiple modalities (visual, acoustic, textual) in videos for personalized recommendation \cite{wei2019mmgcn} and multi-modality biomarkers for accurate diagnosis of Alzheimer's disease \cite{tong2017multi}.
By contrast, there is a stagnation in the EDA domain: 
prior arts that adopt GNNs for fast evaluation focus on mapping static circuit graphs to metrics of interest \cite{zhang2020grannite,wu2022survey,wu2021ironman}.
Thus, the most significant innovation of this work stems from two aspects.
\underline{First}, we consider input information from multiple modalities, i.e., circuit designs in graph format and synthesis flows in sequence format, since the final QoR of circuit designs is dependent on both circuit structures and synthesis flows.
Our investigation with GNN-S and GNN-H shows that efficient approaches to extracting features and fusing information from different modalities can conspicuously improve representation power.
\underline{Second}, we build a large dataset to provide some initial efforts on facilitating multi-modality graph learning in the circuit design and/or EDA domain.
The multi-modal graph representation learning, which integrates the knowledge from other learning schemes with the conventional graph representation learning, is expected to provide more versatility for EDA tasks.

\textbf{Generalization to other transformations.} 
Even though the main focus of this work is the generalization across different circuit designs, which is in fact a more practical case as synthesis tools usually hold a fixed set of transformations awaiting to be applied on different circuit designs \cite{lynx, synopsys, cadence}, we briefly shed light on the feasibility of generalizing to additional transformations.
First, one of the preprocessing steps for LSTM-based models, i.e., the tokenization of logic transformations, includes a special token $<$\texttt{unk}$>$ designed for transformations that are unknown during training yet met in testing.
Second, some out-of-vocabulary techniques in natural language processing (NLP) \cite{hu2019few} can be adopted to improve the generalization capability to new transformations.
Taking a step back, if the introduction of newly developed transformations prompts a need for considerable modifications in the temporal models, either transfer learning specifically tailored for NLP \cite{ruder2019transfer} or complete retraining can be potential solutions.
These training efforts can be recognized as a software update, just as version updates in synthesis tools.

\section{Conclusion}
Aiming to fulfill the two fundamental requirements of production-ready ML in EDA (i.e., the high-accuracy requirement and the generalization capability), we propose a novel approach, LOSTIN, which targets the rapid and accurate performance predictions of logic synthesis flows with great generalization capability across different circuit designs.
We highlight the importance of jointly considering the spatial information from circuit structures and the temporal information from synthesis flows.
Accordingly, two hybrid GNN-based models that simultaneously exploit spatio-temporal information are proposed to predict the synthesized design area and delay.
Specifically, the first model uses a GNN to characterize circuit designs armed with a supernode to encode temporal information, and the second model composes of a GNN for spatial learning and an LSTM for temporal learning.
To evaluate our predictors in both transductive and inductive scenarios, we build a large dataset based on eleven circuit designs.
Evaluation shows that the testing MAPE on designs seen (i.e., transductive) and unseen (i.e., inductive) during training are no more than 1.2\% and 3.1\%, respectively. 
The proposed approach demonstrates great generalization capability across designs, without resorting to any retraining.
\bibliographystyle{IEEEtran}
\bibliography{ref}
\end{document}